\documentclass[letterpaper, 10pt, conference]{ieeeconf}  

\IEEEoverridecommandlockouts                              

\usepackage{graphicx}
\usepackage{xcolor}
\usepackage{amsmath,amsthm,amssymb}
\usepackage[ruled,vlined,linesnumbered]{algorithm2e}
\usepackage[font=footnotesize,labelfont=bf]{caption}    
\usepackage{subcaption}
\usepackage{cite}
\usepackage{url}
\usepackage{comment}
\renewcommand{\tilde}{\widetilde}
\newcommand{\mA}{\mathcal{A}}

\newcommand{\mD}{\mathcal{D}}
\newcommand{\mG}{\mathcal{G}}
\newcommand{\mT}{\mathcal{T}}
\newcommand{\mU}{\mathcal{U}}
\newcommand{\mX}{\mathcal{X}}
\newcommand{\R}{\mathbb{R}}
\newcommand{\Exp}{\mathbb{E}}
\newcommand{\bc}{\mathbf{c}}
\newcommand{\bw}{\mathbf{w}}
\newcommand{\bu}{\mathbf{u}}
\newcommand{\bx}{\mathbf{x}}
\newcommand{\dd}{\mathrm{d}}
\newcommand{\tp}{\mathsf{T}}

\newcommand{\abs}[1]{\left\vert #1 \right\vert}
\newcommand{\norm}[1]{\left\Vert #1 \right\Vert}
\newcommand{\tuple}[1]{\langle #1 \rangle}
\newcommand{\dist}{\operatorname{dist}}
\SetKw{KwInit}{Initialize:}
\SetKw{KwInput}{Input:}
\SetKw{KwOut}{Output:}
\SetKw{KwBreak}{break}
\SetKw{KwOr}{or}

\theoremstyle{remark}

\title{\LARGE \bf
Stackelberg Meta-Learning for Strategic Guidance in Multi-Robot Trajectory Planning
}

\author{Yuhan Zhao and Quanyan Zhu
\thanks{Y. Zhao and Q. Zhu are with the Department of Electrical and Computer Engineering, New York University, Brooklyn, NY, 11201, USA. Email:{\{yhzhao, qz494\}@nyu.edu}.}%
\thanks{This work is accepted by 2023 IEEE/RSJ International Conference on Intelligent Robots and Systems (IROS).}
}

\begin{document}

\maketitle
\thispagestyle{empty}
\pagestyle{empty}

\begin{abstract}
Trajectory guidance requires a leader robotic agent to assist a follower robotic agent to cooperatively reach the target destination. However, planning cooperation becomes difficult when the leader serves a family of different followers and has incomplete information about the followers. There is a need for learning and fast adaptation of different cooperation plans. We develop a Stackelberg meta-learning approach to address this challenge. 
We first formulate the guided trajectory planning problem as a dynamic Stackelberg game to capture the leader-follower interactions. Then, we leverage meta-learning to develop cooperative strategies for different followers. The leader learns a meta-best-response model from a prescribed set of followers. When a specific follower initiates a guidance query, the leader quickly adapts to the follower-specific model with a small amount of learning data and uses it to perform trajectory guidance. 
We use simulations to elaborate that our method provides a better generalization and adaptation performance on learning followers' behavior than other learning approaches. The value and the effectiveness of guidance are also demonstrated by the comparison with zero guidance scenarios\footnote{The simulation codes are available at \url{https://github.com/yuhan16/Stackelberg-Meta-Learning}.}.
\end{abstract}

\section{Introduction} \label{sec:intro}

Guided cooperation is gaining increasing attention in many robotic applications with the advances in robotic research and technology. For example, the path guidance and tracking in multi-robot systems \cite{panagou2015distributed,bibuli2012guidance}, human-robot collaboration in home-assistive services \cite{nikolaidis2017game} and manufacturing \cite{bo2016human}, and multi-robot collective transportation \cite{wang2016kinematic,machado2016multi}.
Guided cooperation can utilize heterogeneous robot capabilities to achieve task objectives. A more resourceful robotic agent (leader) can guide or assist a less sophisticated robotic agent (follower) to complete the task by utilizing both agents' advantages. A typical and important application of guided cooperation in robotics is strategic guidance for trajectory planning. As an example, we consider an unmanned aerial vehicle (UAV) guiding an unmanned ground vehicle (UGV) for transporting mission-critical objects. 
The UGV has limited sensing and computational resources and can only do its local planning, resulting in difficulty reaching the destination independently. In contrast, the resourceful UAV can sense the global environment and find a collision-free trajectory to guide the UGV.

Extensive works in trajectory planning, including path-based planning \cite{kavraki1996probabilistic,lavalle2001randomized,karaman2011sampling} and control-based planning \cite{wang2009fast,mohanan2018survey}, have investigated single-agent cases. However, simply extending the methodology to multi-agent settings is insufficient since it fails to capture the cooperative interactions between multiple agents. Game theory emerges as a promising tool in multi-robot trajectory planning \cite{hang2020human,wang2021game,turnwald2019human,zhu2014game,wang2020game}.
In particular, Stackelberg game \cite{bacsar1998dynamic} provides a quantitative framework that captures the mentor-apprentice or leader-follower type of interactions in guided cooperation and has shown effectiveness in various cooperative tasks \cite{koh2020cooperative,sadigh2016planning,fisac2019hierarchical,zhao2022stackelberg}. In the UAV-UGV example, interactions between UAV and UGV in trajectory guidance can be formulated as a dynamic Stackelberg game. The associated Stackelberg equilibrium solution provides an agent-wise cooperative plan.

\begin{figure}
    \centering
    \includegraphics[scale=0.17]{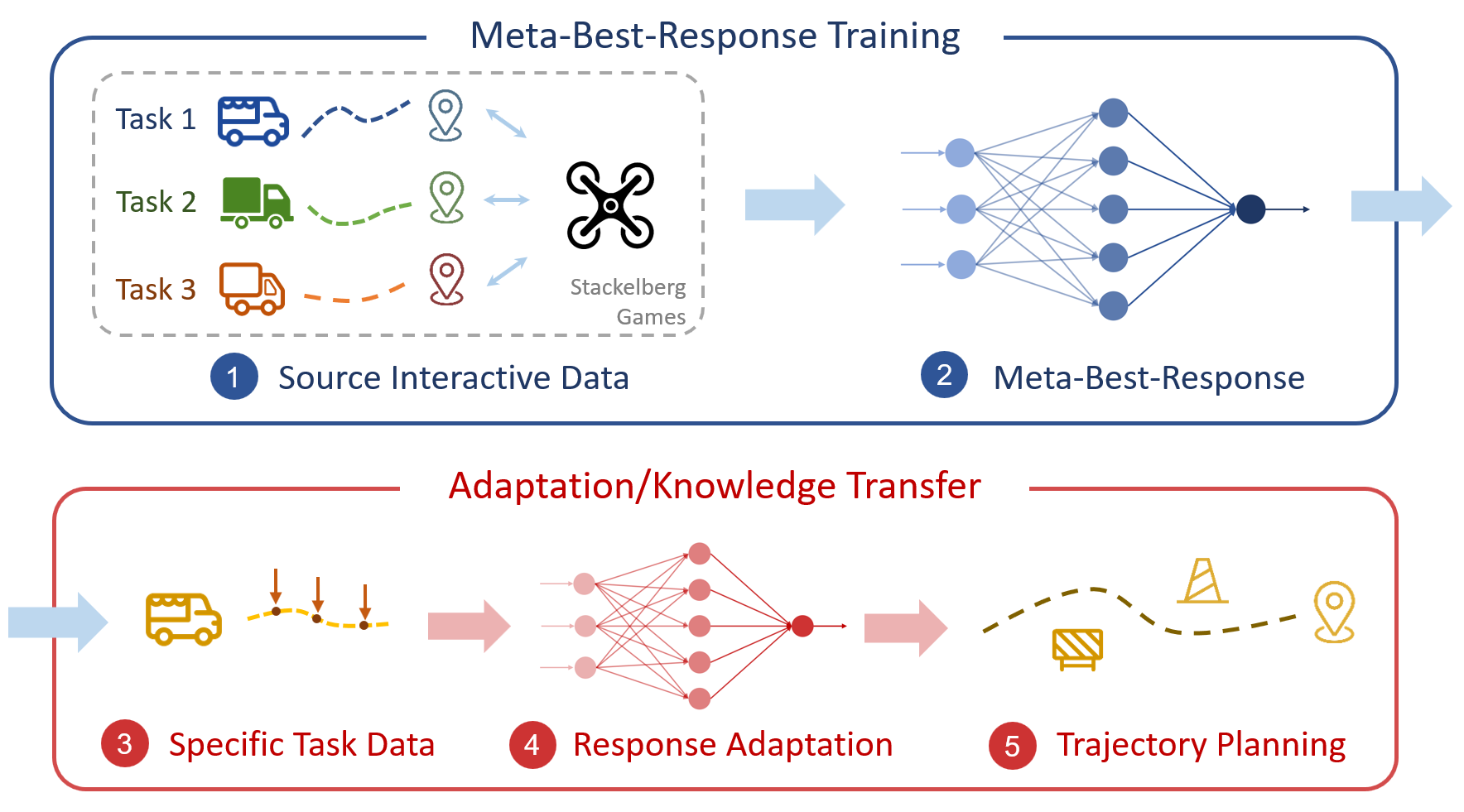}
    \caption{Illustration of Stackelberg meta-learning approach in trajectory guidance. Different follower UGVs rely on the leader UAV's trajectory guidance to reach their destinations. The leader UAV uses meta-learning to learn a meta-best-response model by interacting with different followers (1-2). When guiding a specific follower, the leader uses the follower-specific data (3) to adapt the meta-model to that follower (4) and performs guided trajectory planning (5).
    }
    \label{fig:intro}
\end{figure}

However, the leader must know the follower's behavior model to solve the Stackelberg equilibrium, which is challenging in practical applications where the leader can only observe the follower's action instead of his model. Besides, a leader generally needs to work with different followers on various tasks. In the trajectory guidance example, a UAV is expected to assist UGVs of different configurations to accomplish the transport mission. Designing different cooperation plans for various followers can be time-consuming and eventually becomes intractable as the number of followers increases. We need an approach to quickly generate a guidance plan when the leader assists a specific follower.

Meta-learning provides a suitable learning mechanism for Stackelberg game-theoretic cooperation (see Fig.~\ref{fig:intro}). It enables learning a customizable plan from a prescribed set of known tasks and fast adaptation to a specific task using a small amount of learning data \cite{finn2017model}. 
It has also been used in robotics to search for adaptive collaboration plans in multi-robot systems \cite{Jia2022crmrl}, human-robot interaction \cite{gao2019fast,luipers2021concept}, and trajectory planning and tracking \cite{xu2022meta,richards2022control}.
Specifically, the leader prepares a meta-best-response for different followers as an averaged behavior model based on past interactions. When a specific follower initiates a guidance query, the leader adapts the meta-model to the follower and then performs trajectory guidance by finding an approximate Stackelberg equilibrium. 

In this work, we develop a Stackelberg meta-learning approach for leader-follower cooperation in guided trajectory planning problems. We formulate the guided interactions between the leader and the follower as a dynamic Stackelberg game, where each follower makes \emph{myopic} decisions to interact with the leader. Once a specific follower initiates a guidance query, the leader quickly generates a follower-specific behavior model from the meta-best-response and uses it to design effective trajectory guidance strategies with receding horizon planning. 
We use simulations to show that our approach provides a better generalization and adaptation performance compared with the other two typical learning approaches. We also compare the results with zero guidance scenarios to demonstrate the value of guidance in trajectory planning.

\emph{Notations:} We use superscripts $L$ and $F$ to denote the leader and the follower-related quantities. We use bold variables to denote aggregated trajectory, e.g., $\bu := \{u_t\}_{t=0}^{T-1}$. We use the two-norm $\norm{x}_2 = \sqrt{x^\tp x}$ and the matrix norm $\norm{x}_Q = \sqrt{x^\tp Q x}$.

\section{Problem Formulation}
\subsection{Trajectory Guidance as Stackelberg Games}
We consider a leader robotic agent $L$ (she) assisting the follower robotic agent $F$ (he) to reach the destination while circumventing obstacles in the working environment $\mX \subset \R^2$. Due to limited sensing and computational capabilities, the follower requires guidance from the leader to find a collision-free trajectory.
The leader is expected to assist different followers in trajectory planning. The followers with different configurations are characterized by their type $\theta \in \Theta$, and $p(\theta)$ is the type distribution known to the leader. The leader assists one follower at a time when a specific follower drawn from $p(\theta)$ requests a guidance query.

Let $x^L_t \in \R^{n^L}, x^F_t \in \R^{n^F}$ denote the leader and follower's states, including their positions at time $t$; $x_t := [x^L_t, x^F_t]$ denotes the joint state. Let $u^L_t \in \mU^L \subset \R^{m^L}, u^F_t \in \mU^F \subset \R^{m^F}$ be their controls and admissible control sets. 
In the guidance of a follower with type $\theta$, the leader announces an action $u^L_t$ based on $x_t$. The follower then myopically responds to the leader with the optimal action $u^{F*}_\theta(x_t, u^L_t)$ based on his cost function $J^F_\theta$. Using the follower's response, the leader seeks a collision-free trajectory $\bx^* :=\{x^*_t\}_{t=0}^T$ for both agents and strategically guides the follower to the destination. We formulate the trajectory guidance problem as a dynamic Stackelberg game $\mG_\theta$ as follows:
\begin{align}
    \min_{u^L \in \mU^L} \ & J^L_\theta(\bu^L) := \sum_{t=0}^{T-1} g^L_\theta(x_t, u^L_t, u^{F*}_\theta(x_t, u^L_t)) + q^L_\theta(x_T)  \label{eq:sg.obj} \\
    \text{s.t.} \ & u^{F*}_\theta(x_t, u^L_t) = \arg\min_{u^F \in \mU^F} \ J^F_\theta(x_t, u^L_t, u^F_t), \notag \\ 
    &\hspace{4cm} t=0, \dots, T-1, \label{eq:sg.br} \\ 
    & x^L_{t+1} = f^L(x^L_t, u^L_t), \quad t=0, \dots, T-1, \label{eq:sg.dynamics_leader}\\
    & x^F_{t+1} = f^F_\theta(x_t, u^L_t, u^F_t), \quad t=0,\dots, T-1, \label{eq:sg.dynamics_follower} \\ 
    &\dist(x^i_t, p^O_j) \geq d_j, \quad i \in \{L,F\}, \ j = 1,\dots, M, \notag \\
    &\hspace{4cm} t=0,\dots, T. \label{eq:sg.safe} 
\end{align}
Here, $g^L_\theta$ and $q^L_\theta$ denote the leader's stage and terminal costs. The follower's problem is given by \eqref{eq:sg.br} where $J^F_\theta$ is his cost function. 
The leader and follower's dynamic models are captured by \eqref{eq:sg.dynamics_leader} and \eqref{eq:sg.dynamics_follower}, respectively.
In the safety constraints \eqref{eq:sg.safe}, $p^O_j \in \mX$ is the position of the $j$-th obstacle in $\mX$ and $d_j$ is the safety distance. The function $\dist$ measures the distance between the leader/follower and an obstacle.

We use the open-loop Stackelberg equilibrium $\tuple{\bx^*, \bu^{L*}, \bu^{F*}_\theta}$ of the game $\mG_\theta$ as the cooperative plan for trajectory guidance. The leader and the follower take the actions in $\bu^{L*}$ and $\bu^{F*}_\theta$ respectively and generates a collision-free trajectory starting from given $x_0 := [ x^L_0, x^F_0 ]$.

\subsubsection{Guidance in Trajectory Planning}
Guidance is an interactive process instead of unilateral instructions from the leader. Followers have the freedom to (myopically) decide where to go after observing the state $x_t$ and the leader's action $u^L_t$. The leader's trajectory serves as a reference to assist followers to perform effective local planning. When there is no guidance, followers can only perform one-step planning
using limited sensing and planning capabilities.
It is worth noting that the leader's recommended action coincides with the follower's optimal action $u^{F*}_\theta$ only if the leader knows the follower's exact decision-making model. In this case, followers follow the recommendation and do not need local planning. However, 
the anticipated action may not be taken if the leader only has an approximate follower model. Then all followers need to make decisions by themselves in the guidance task.

\subsection{Meta-Best-Response and Meta-Learning Problem}
When the leader knows the follower's decision-making model ($J^F_\theta$ or $u^{F*}_\theta$), 
the Stackelberg equilibrium can be computed using model-based approaches, such as mixed integer linear programming \cite{paruchuri2008playing,zhao2022stackelberg}. However, the leader can only observe the follower's actions in many practical cases. We need learning-based methods to find the Stackelberg equilibrium of $\mG_\theta$ to design effective guidance plans.

We approximate the follower's decision problem \eqref{eq:sg.br} using an input-output function (i.e., the best-response model) $b(x, u^L; \bw)$ parameterized by $\bw$. The leader uses $b$ to predict the follower's response and $\mG_\theta$ becomes a parameterized single-agent trajectory optimization problem, denoted by $\mG_\theta(\bw)$ for a given parameter $\bw$.
The optimal solution $\{ \bx^*(\bw), \bu^{L*}(\bw) \}$ of $\mG_\theta(\bw)$ together with the estimated best response $\{ b(x^*_t(\bw), u^{L*}_t(\bw) ) \}_{t=1}^{T-1}$ approximate the Stackelberg equilibrium of $\mG_\theta$ and is used for trajectory guidance.

When the leader assists different followers, it is time-consuming for the leader to learn separate best-response models from scratch. We aim to develop an approach to quickly generate a guidance plan when a follower initiates a query. To this end, we formulate a meta-learning problem such that the leader learns a meta-best-response model from a set of sampled or encountered followers through offline learning. When a new query occurs, the leader uses a small amount of learning data (from history) to adapt the meta-best-response model to the follower-specific one and performs trajectory guidance.

With a little abuse of notation, we denote $b(x, u^L; \bw)$ as the meta-best-response model and define task $\mT_\theta$ as learning the best-response of the type $\theta$ follower with the task cost
\begin{equation}
    \label{eq:fitting_cost}
    L_\theta(\bw) = \frac{1}{N} \sum_{i=1}^N \norm{b(\hat{x}_i,\hat{u}^L_i; \bw) - \hat{u}^{F*}_{\theta,i}}_2^2,
\end{equation}
where $\mD_\theta = \{ (\hat{x}_i, \hat{u}^L_i, \hat{u}^{F*}_{\theta,i}) \}_{i=1}^N$ is the sampled best-response data. $L_\theta(\bw)$ measures the data fitting cost over $\mD_\theta$. The meta-learning can be formulated as the following optimization problem:
\begin{equation}
\label{eq:meta.1}
    \min_\bw \quad \Exp_{\theta \sim p} \left[ \Exp_{\mD_\theta}[L_\theta(\bw - \alpha \nabla_\bw L_\theta(\bw))] \right].
\end{equation}
where $\alpha > 0$ is the inner gradient update step size.

\section{Stackelberg Meta-Learning} \label{sec:sgmeta}

\subsection{Meta-Best-Response Training}
We use the empirical task distribution to approximate the expectation in \eqref{eq:meta.1} and obtain
\begin{equation}
\label{eq:meta.2}
    \min_{\bw} \quad
    \sum_{\theta \sim p} \sum_{\mD_\theta} L_\theta \left( \bw - \alpha \nabla_\bw L_\theta(\bw; \mD^{\text{train}}_\theta); \mD^{\text{test}}_\theta \right).
\end{equation}
Here, we split the dataset $\mD_\theta = \mD^{\text{train}}_\theta \cup \mD^{\text{test}}_\theta$; $\theta \sim p$ is the empirical task distribution of sampled batch tasks $\mT_{\text{batch}}$ from $p$.
We use first-order methods to solve \eqref{eq:meta.2}. The intermediate parameter $\bw'_\theta$ for task $\mT_\theta$ is updated by one gradient step:
\begin{equation}
\label{eq:meta-param.1}
    \bw'_\theta \gets \bw_k - \alpha \nabla_\bw L_\theta(\bw_k; \mD^{\text{train}}_\theta). 
\end{equation}
Next, we perform stochastic gradient descent (SGD) on the meta-optimization across $\mT_{\text{batch}}$ with the step size $\beta > 0$:
\begin{equation}
\label{eq:meta-param.2}
    \bw_{k+1} \gets \bw_k - \frac{\beta}{\abs{\mT_{\text{batch}}}} \sum_{\theta \in \mT_{\text{batch}}} \nabla_\bw  L_\theta(\bw'_\theta; \mD^{\text{test}}_\theta),
\end{equation}
Alg.~\ref{alg:meta-learning} summarizes the Stackelberg meta-learning algorithm and outputs a meta-best-response model as the averaged behavior model for all types of followers.

\begin{algorithm}
\KwInput Follower's type distribution $p(\theta)$, hyperparameters $\alpha, \beta$,  $\mathrm{MAX\_ITER}$ \;
\KwInit Initial model parameter $\bw_0$, \;
$k \gets 0$ \;
\While{$k < \mathrm{MAX\_ITER}$}{
    Sample a batch of tasks $\mT_{\text{batch}}: = \{ \mT_\theta \}, \theta \sim p$ \;
    \tcp{Inner level gradient evaluation}
    \For{every task $\mT_\theta \in \mT_{\mathrm{batch}}$}{
        Sample $K_1$ best response data given random feasible state-action pair $(x,u^L)$ \;
        Sample $K_2$ best response data near the obstacle $j = 1, \dots, M$ \;
        $\mD^{\text{train}}_\theta \gets$ all samples with $K=K_1+K_2$ \;
        Compute $\bw'_\theta$ by \eqref{eq:meta-param.1} and $\mD^{\text{train}}_\theta$ \;
        $\mD^{\text{test}}_\theta \gets$ sample $K$ best response data using $\bw'_\theta$, following same sample rule as $\mD_\theta^{\text{train}}$ \;
    }
    \tcp{Meta-optimization evaluation}
    Update $\bw_{k+1}$ by \eqref{eq:meta-param.2} and $\{ \mD_\theta^{\text{test}} \}_{\theta \sim p}$ \;
    $k \gets k + 1$ \;
}
$\bw_\mathrm{meta} \gets \bw_k$ \;
\KwOut meta-parameter $\bw_\mathrm{meta}$ and meta-best-response model $b(x,u^L; \bw_\mathrm{meta})$ \;
\caption{Stackelberg meta-learning algorithm.}
\label{alg:meta-learning}
\end{algorithm}

\subsubsection{Importance Sampling}
The follower's action near an obstacle differs from that in the flat region. The best-response model should capture the follower's response in both situations as precisely as possible. Using the idea of importance sampling, we randomly sample $K_1$ data in $\mX$ and sample $K_2$ data near the obstacles and use $\kappa := K_1 / K_2$ to control the sampling ratio.

\subsection{Best-Response Adaption}
After receiving $b(x,u^L; \bw_{\mathrm{meta}})$ from Alg.~\ref{alg:meta-learning}, the leader can fast adapt to the new task (the new follower) using only a small amount of data samples when a guidance query occurs. Specifically, the leader adapts $b(x,u^L; \bw_{\mathrm{meta}})$ to 
a follower-specific best-response model $b(x,u^L; \bw_\theta)$ using the adaption algorithm Alg.~\ref{alg:meta-adaption} and uses it to perform trajectory guidance.

\begin{algorithm}
\KwInput $\bw_{\mathrm{meta}}$ and new follower with type $\theta$ \;
\KwInit Step size parameter $\alpha$, adaption step $C$ \;
Sample $\mD_\theta$ with $K'$ samples using $\bw_{\mathrm{meta}}$ \;
$k \gets 0$, $\bw_0 \gets \bw_{\mathrm{meta}}$ \;
\While{$k < C$}{
    $\bw_{k+1} \gets \bw_k - \alpha \nabla_w L_\theta(\bw_k)$ with $\mD_\theta$ \;
    $k \gets k + 1$ \;
}
Adapted parameter $\bw_\theta \gets \bw_k$ \;
\KwOut adapted parameter $\bw_\theta$ and adapted best response model $b(x,u^L; \bw_\theta)$ \;
\caption{Adaption of Stackelberg meta-learning.}
\label{alg:meta-adaption}
\end{algorithm}

\subsection{Receding Horizon Planning For Trajectory Guidance}
The leader uses the adapted best-response model and solves the problem $\mG_\theta(\bw)$ to perform trajectory guidance. 
The safety constraints \eqref{eq:sg.safe} bring computational challenges. We penalize the violation of safety constraints using barrier functions for $j=1,\dots,M$:
\begin{equation*}
    c_{j}(x_t) = - \sum_{i \in \{L,F\}} \nu \log \left( \dist(x^i_t, p^O_j) - d_j \right),
\end{equation*}
where $\nu > 0$ is the penalty parameter. Then, the modified cost function of $\mG_\theta(\bw)$ is given by
\begin{equation*}
\begin{split}
    &\tilde{J}^L_\theta(\bu^L; \bw) := \\
    \sum_{t=0}^{T-1} & \left[ \tilde{g}^L_\theta(x_t, u^L_t; \bw) + \sum_{j=1}^M c_{j}(x_{t}) \right] + q^L_\theta(x_T) + \sum_{j=1}^M c_j(x^i_T),
\end{split}
\end{equation*}
where $\tilde{g}^L_\theta(x_t, u^L_t; \bw) := g^L_\theta(x_t, u^L_t, b(x_t, u^L_t; \bw))$. Instead of solving $\mG_\theta(\bw)$. We solve the following modified problem
\begin{equation*}
\begin{split}
    &\tilde{\mG}_\theta(\bw): \quad \min_{\bu^L \in \mU^L} \ \tilde{J}^L_\theta(\bu^L; \bw) \\
    &\hspace{6mm} \text{s.t.} \ x^L_{t+1} = f^L(x^L_t, u^L_t), \ t=0\dots,T-1, \\ 
    &\hspace{11.5mm} x^F_{t+1} = f^F_\theta(x_t, u^L_t, b(x_t, u^L_t; \bw)), \ t=0,\dots, T-1,
\end{split}
\end{equation*}
and uses its solution $\{ \tilde{\bx}^*(\bw), \tilde{\bu}^{L*}(\bw) \}$ to approximate the one of $\mG_\theta(\bw)$.
We can leverage existing optimization solvers to solve $\tilde{\mG}_\theta(\bw)$. However, we note that the follower's dynamics $f^F_\theta$ contain the parameterized best-response model $b(x,u^L; \bw)$. Parameterized models in general have complex function surfaces, such as neural networks \cite{lecun2015deep}. The direct use of the parameterized model in the equality constraint may result in infeasible solutions. 
To address this issue, we penalize the follower's dynamics difference by putting it into the objective function. We define
\begin{equation*}
    d_t(x_{t+1}, x_t, u^L_t; \bw) = \mu \norm{x^F_{t+1} - x^F_t - f^F_\theta(x_t, u^L_t; \bw)}_2^2
\end{equation*}
for $t=0,\dots, T-1$, where $\mu > 0$ is the penalty parameter. Hence we solve the following problem 
\begin{equation*}
\begin{split}
    \tilde{\mG}^{\mathrm{opt}}_\theta(\bw): \ \ & \min_{\bu^L \in \mA} \quad \tilde{J}^L_\theta(\bu^L; \bw) + \sum_{t=0}^{T-1} d_t(x_t, x_t, u^L_t; \bw), \\
    &\text{s.t.} \quad x^L_{t+1} = f^L(x^L_t, u^L_t), \quad t=0\dots,T-1,
\end{split}
\end{equation*}
to obtain an approximate solution. We further use the necessary optimality conditions of $\tilde{\mG}_\theta(\bw)$ derived from Pontryagin's Minimum Principle (PMP) to refine the approximate solution of $\tilde{\mG}^{\mathrm{opt}}_\theta(\bw)$:
\begin{equation}
\label{eq:pmp}
\begin{split}
    x_{t+1} &= f(x_t, u^L_t; \bw), \quad t=0,\dots, T-1, \\ 
    \lambda_t &= \nabla_{x_t} H_t(x_t, u^L_t, \lambda_{t+1}), \quad t=1,\dots, T-1,\\ 
    \lambda_T &= \nabla_{x_T} q^L_\theta(x_T) + \nabla_{x_T} \sum_{j=1}^M c_j(x_T), \\ 
    u^L_t &= \arg\min_{u \in \mU^L} H_t(x_t, u, \lambda_{t+1}), \ t=0,\dots, T-1,
\end{split}    
\end{equation}
where $f(x_t, u^L_t; \bw) := [f^L(x_t, u^L_t), f^F_\theta(x_t, u^L_t, b(x_t, u^L_t; \bw))]$ is the aggregated dynamics, $\lambda_t \in \R^{n^A+n^B}$ is the costate at time $t$, and the Hamiltonian $H_t$ is given by
\begin{equation*}
\begin{split}
    H_t(x_t, u^L_t, \lambda_{t+1}) := \\ \tilde{g}^L_\theta(x_t, u^L_t; \bw) &+ \sum_{j=1}^M c_j(x_t) + \lambda^\tp_{t+1} f(x_t,u^L_t; \bw)
\end{split}
\end{equation*}
for $t=0,\dots, T-1$. 
Gradient methods can be applied to find the minimizer of $H_t$ in \eqref{eq:pmp}.

Based on $\tilde{\mG}^{\mathrm{opt}}_\theta$ and \eqref{eq:pmp}, we use receding horizon planning to generate effective and robust trajectory guidance strategies, which is summarized in Alg.~\ref{alg:receding-horizon}.

\begin{algorithm}
\KwInput Query type $\theta \in \Theta$, initial state $x_{\mathrm{init}}$, destination $p^\dd$, $\mathrm{MAX\_TIME}$ \;
Run Alg.~\ref{alg:meta-adaption} to obtain the adapted model $b(x,a; \bw_\theta)$\;
$x^\dd \gets$ form target state \; 
$t \gets 0$, $x_t \gets x_{\mathrm{init}}$ \;
\While{True}{
    Leader sets $x_t$ as the initial state in $\tilde{\mG}^{\mathrm{opt}}_\theta(\bw)$ \;
    $\bar{\bu}^L$ $\gets$ Leader solves $\tilde{\mG}^{\mathrm{opt}}_\theta(\bw)$ \;
    $\tilde{\bu}^{L*}(\bw)$ $\gets$ Leader refines $\bar{\bu}^L$ by solving \eqref{eq:pmp} \;
    Leader announces $\tilde{u}^{L*}_t(\bw)$ to the follower \;
    $u^{F*}_t \gets$ follower observes $x_t$ and $\tilde{u}^{L*}_t$ \;
    $x_{t+1} \gets$ real system dynamics \eqref{eq:sg.dynamics_leader}-\eqref{eq:sg.dynamics_follower} \;
    
    \uIf{Reach destination \KwOr $t > \mathrm{MAX\_TIME}$}{
        \KwBreak
    }
    $x_t \gets x_{t+1}$\;$t \gets t+1$ \;   
}
\caption{Receding horizon planning.}
\label{alg:receding-horizon}
\end{algorithm}

\section{Simulations and Evaluations} \label{sec:exp}

\subsection{Simulation Settings}
We choose a $[0,10] \times [0,10]$ working space $\mX$ with four obstacles shown in Fig.~\ref{fig:rc_trajectory}. We use a single integrator $\dot{p}^L = v^L$ as the leader's dynamic model, where $p^L, v^L \in \R^2$ are the leader's position and velocity. We use a unicycle model to all followers: $\dot{\phi}^F = \omega^F, \dot{p}^F_x = v^F \cos(\phi^F), \dot{p}^F_y = v^F \sin(\phi^F)$, where $\phi^F \in (-\pi, \pi], p^F := [p^F_x, p^F_y] \in \R^2$ denotes the rotation angle and $x,y$-positions. $\omega^F, v^F \in [-1,1],$ are the input angular velocity and linear velocity. We assume that the leader and all followers know the dynamic models.
We denote the joint state $x := [ p^L, p^F, \phi^F ]$, $u^L := v^L$, and $u^F := [ v^F, \omega^F ]$. The interaction (planning) time is set as $T = 2$s, and the discretization time step is $\dd t = 0.2$s. 
The discrete-time model is discussed in Appendix~\ref{app:dynamic_model}. 
We assume that all followers have the same destination $p^\dd = [9,9]$. For simplicity, we elaborate on the definitions of the leader and followers' cost functions in Appendix~\ref{app:cost_fn}. 
The penalty parameters $\nu = 0.5$ and $\mu = 50$. 

We consider five types of followers ($\abs{\Theta}=5$) with a type distribution $p = [0.2, 0.3, 0.1, 0.3, 0.1]$. Using the cost function definition \eqref{eq:follower_cost}, 
the follower in each type has a different set of parameters $\bc := [c_1,\dots,c_4]$. Type 1: $\bc=[1, 8, 1, 0.8]$. Type 2: $\bc = [1, 10, 2, 0.7]$. Type 3: $\bc = [1, 10, 2, 0.6]$. Type 4: $\bc = [1, 5, 0.5, 1]$. Type 5: $\bc = [1, 5, 0.3, 1.2]$. Intuitively, we can label type 2-3 as ``careful" agents and type 4-5 as ``aggressive" agents. This is because type 2-3 followers are more sensitive to the guidance cost and have a wider sensing range. They tend to follow the leader more closely than type 4-5 followers.

We note from \eqref{eq:sg.br} that collecting the follower's best response data does not need the leader to take optimal actions. The leader can record the follower's response by providing a feasible state $x$ and action $u^L$ and prepare the dataset.

\subsection{Meta-Learning Results}
We use a neural network with two hidden layers of 50 ReLU nonlinearities to parameterize the follower's best response. We perform $5\times10^5$ iterations for meta-optimization. In each meta iteration, we sample 5 tasks ($\abs{\mT_{\text{batch}}}= 5$) and $K=100$ for $\mD^{\text{train}}_\theta$ and $\mD^{\text{test}}_\theta$. The sample ratio $\kappa = 2$. In adaptation, we sample $K'=1000$ data with the same $\kappa$. 

Since meta-training aims to find an averaged initial model for fast adaptation, we implement another two model-averaging approaches for comparison. We first train a model to average the \emph{output space} (Output-Ave), i.e., training a best-response model with the same structure as meta-learning using the shuffled data of types of followers (average followers' behaviors). We also train a model to average the \emph{parameter space} (Param-Ave), i.e., training $\abs{\Theta}$ individual models for each type of follower and then averaging the model parameters using $p(\theta)$. Each individual model is trained by supervised learning and the type-specific dataset. 
These two comparative approaches are typical ways to generate averaged models used for adaptation. They have trained over $10^4$ epochs with SGD with fine-tuned hyperparameters. Each epoch contains $150$ iterations. All learning algorithms are performed on AMD Ryzen 3990X CPU. Tab.~\ref{tab:learning_data} summarizes the data usage and training time of different approaches. We manually implement the training of Output-Ave and Param-Ave by PyTorch but implement meta-learning algorithms, which explains the training time difference.

\begin{table}[h]
    \centering
    \begin{tabular}{c|c|c|c|c} \hline
        & Meta-learning& Output-Ave & Param-Ave & Adaptation \\ \hline
        Time & $80.6$ min & $24.2$ min & $27.4$ min (of 5) & $5.3$ s \\ \hline
        Data & $7.5\times 10^4$ (of 5)& $1.5\times 10^4$ & $7.5\times 10^4$ (of 5) & $1000$ \\ \hline
    \end{tabular}
    \caption{Summary of learning statistics. \emph{Output-Ave} means averaging the output space; \emph{Param-Ave} means averaging the parameter space. Training time for Param-Ave is averaged on separate models. The data for meta-learning and Param-Ave are summed over types.}
    \label{tab:learning_data}
\end{table}

\begin{figure}
    \centering
    \begin{subfigure}[t]{0.23\textwidth}
        \centering
        \includegraphics[height=3.12cm]{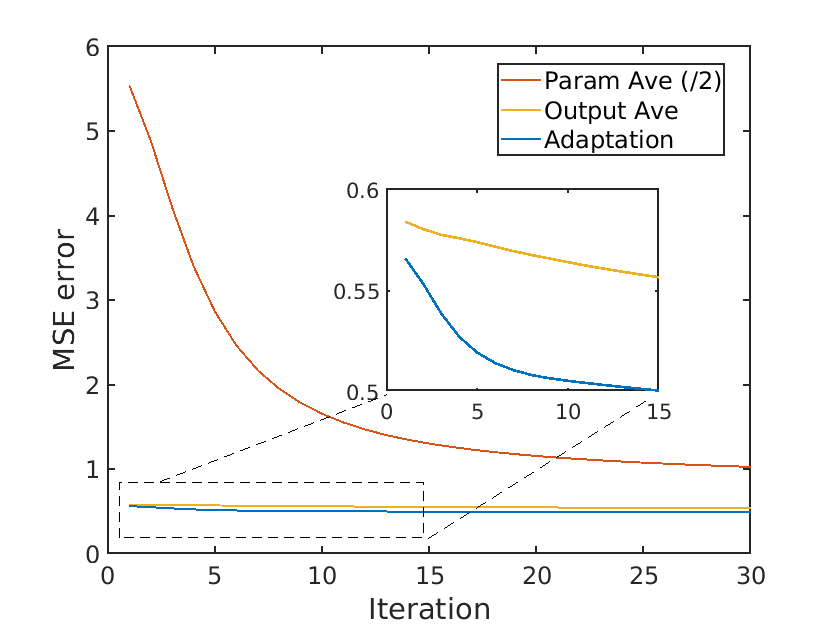}
        \caption{Adaptation loss curve for type 2 follower. A zoom-in box shows more details of meta adaptation and Output-Ave results.}
        \label{fig:meta_adapt.1}
    \end{subfigure}
    \hspace{2mm}
    \begin{subfigure}[t]{0.23\textwidth}
        \centering
        \includegraphics[height=3.3cm]{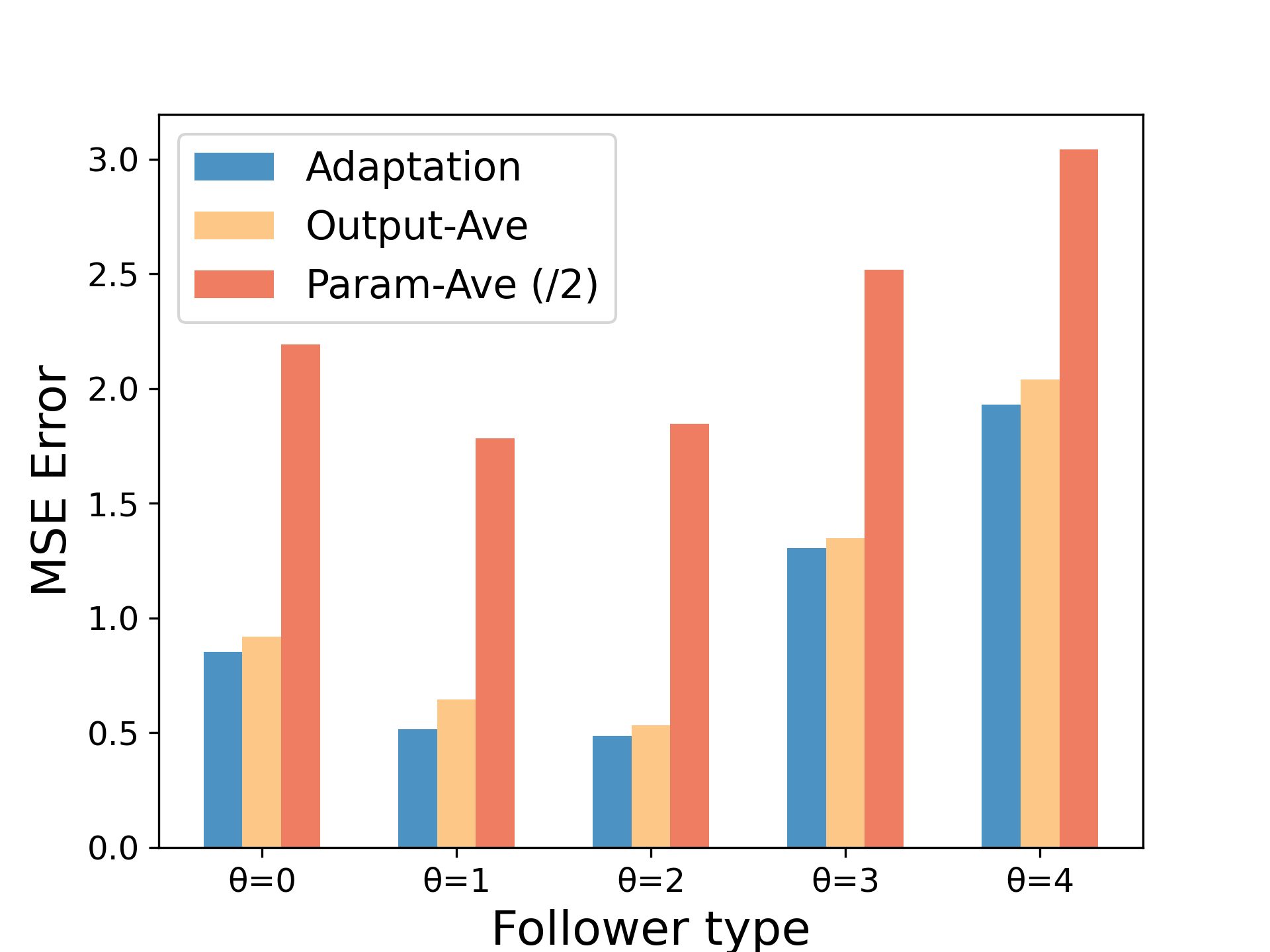}
        \caption{MSE errors after $C=50$ gradient steps adaptation for all followers. Meta-learning provides the best adaptation result.}
        \label{fig:meta_adapt.2}
    \end{subfigure}
    \caption{Adaptation results for three learning approaches. Meta-learning provides the best generalization adaptation performance. Param-Ave approach yields a significant adaptation error and poor generalization performance. We divide its loss by 2 in both plots for better visualization.}
    \label{fig:meta_adapt}
\end{figure}

We perform $C=50$ steps of gradient adaptation for three approaches using the same sampled dataset with $\alpha = 10^{-4}$. For a detailed view, we plot the adaptation loss curve for type $\theta=2$ follower in Fig.~\ref{fig:meta_adapt.1} and use the bar plot to show the adapted result (MSE loss) for all type $\theta$ in Fig.~\ref{fig:meta_adapt.2}. 

The smaller adaptation error and the faster convergence rate indicate a better-generalized performance. As we observe in Fig.~\ref{fig:meta_adapt.1}, using the meta-learned model can fast reduce the adaptation error in the first few gradient steps compared with the Output-Ave approach, meaning that the meta-model can be adapted to the specific follower using fewer amounts of data and generating better performances. The meta-learned model performs best after $C$ adaptation rounds, as shown in Fig.~\ref{fig:meta_adapt.2}. Conversely, the significant adaptation error produced by Param-Ave shows that averaging the model is not a practical option for predicting a new follower's behavior in real-world applications.

\subsection{Receding Horizon Planning}
\begin{figure*}
    \captionsetup[subfigure]{justification=centering}
    \begin{subfigure}[b]{0.19\textwidth}
        \centering
        \includegraphics[height=3.5cm]{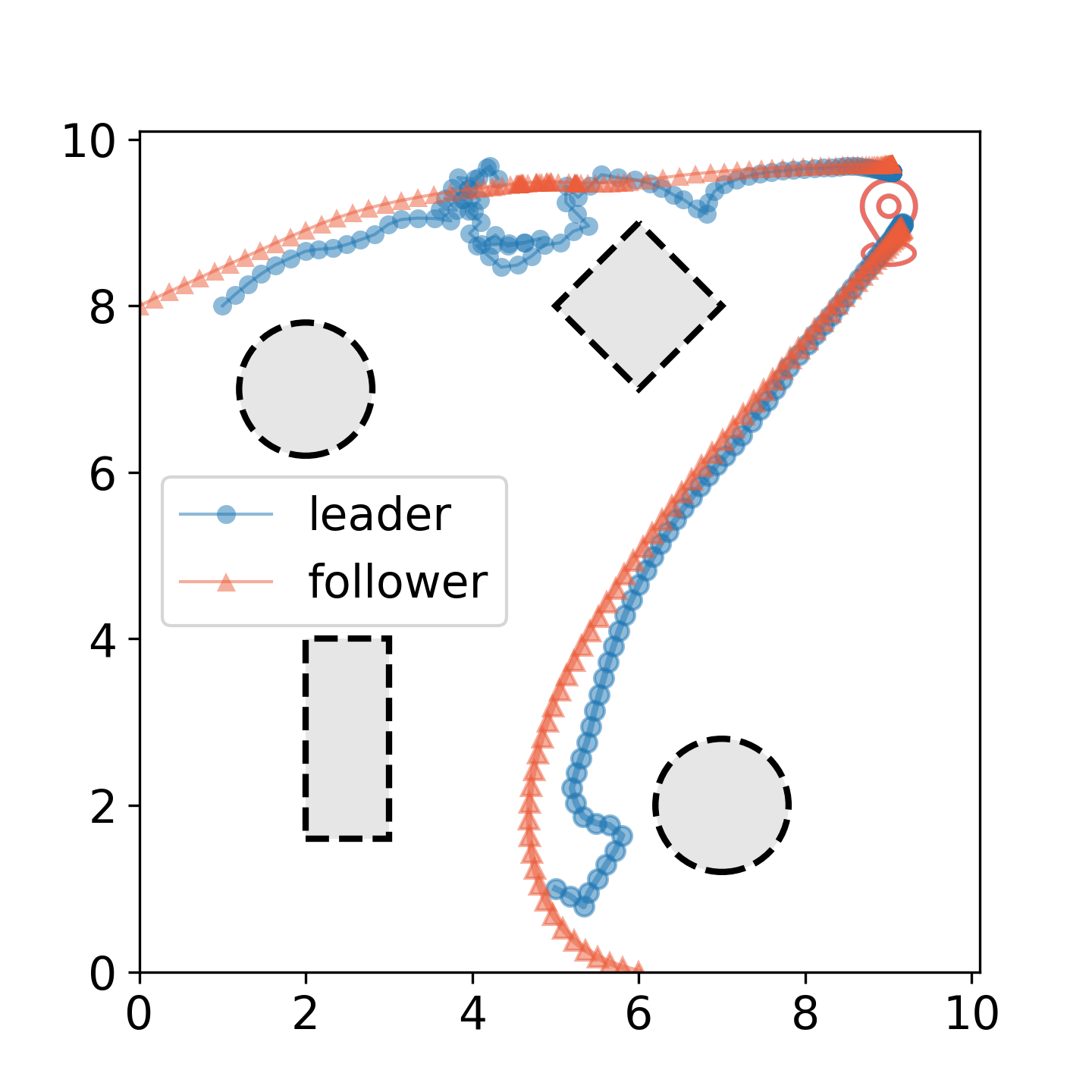}
        \caption{Type 1 follower.}
        \label{fig:type.1}
    \end{subfigure}
    \captionsetup[subfigure]{justification=centering}
    \begin{subfigure}[b]{0.19\textwidth}
        \centering
        \includegraphics[height=3.5cm]{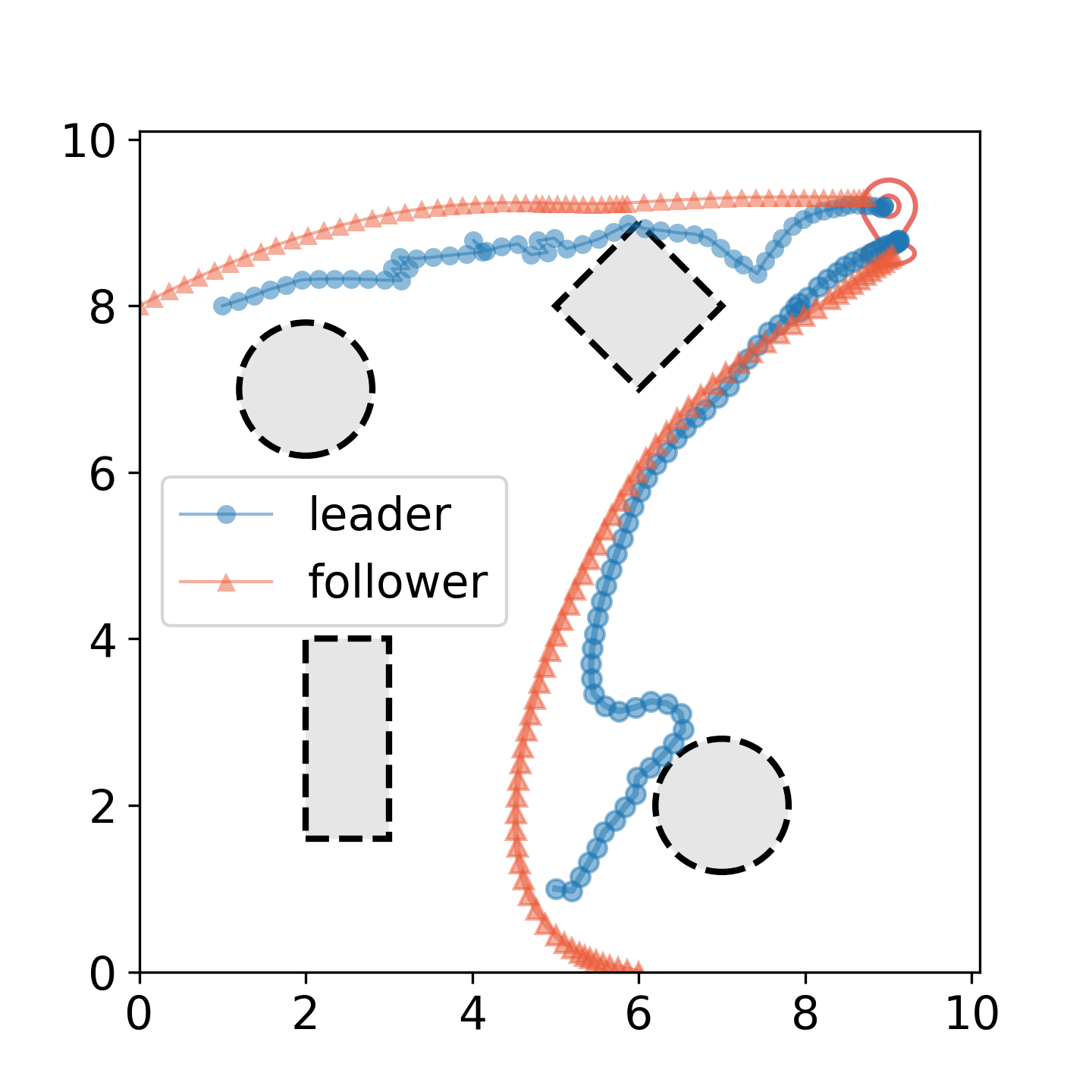}
        \caption{Type 2 follower.} 
        \label{fig:type.2}
    \end{subfigure}
    \captionsetup[subfigure]{justification=centering}
    \begin{subfigure}[b]{0.19\textwidth}
        \centering
        \includegraphics[height=3.5cm]{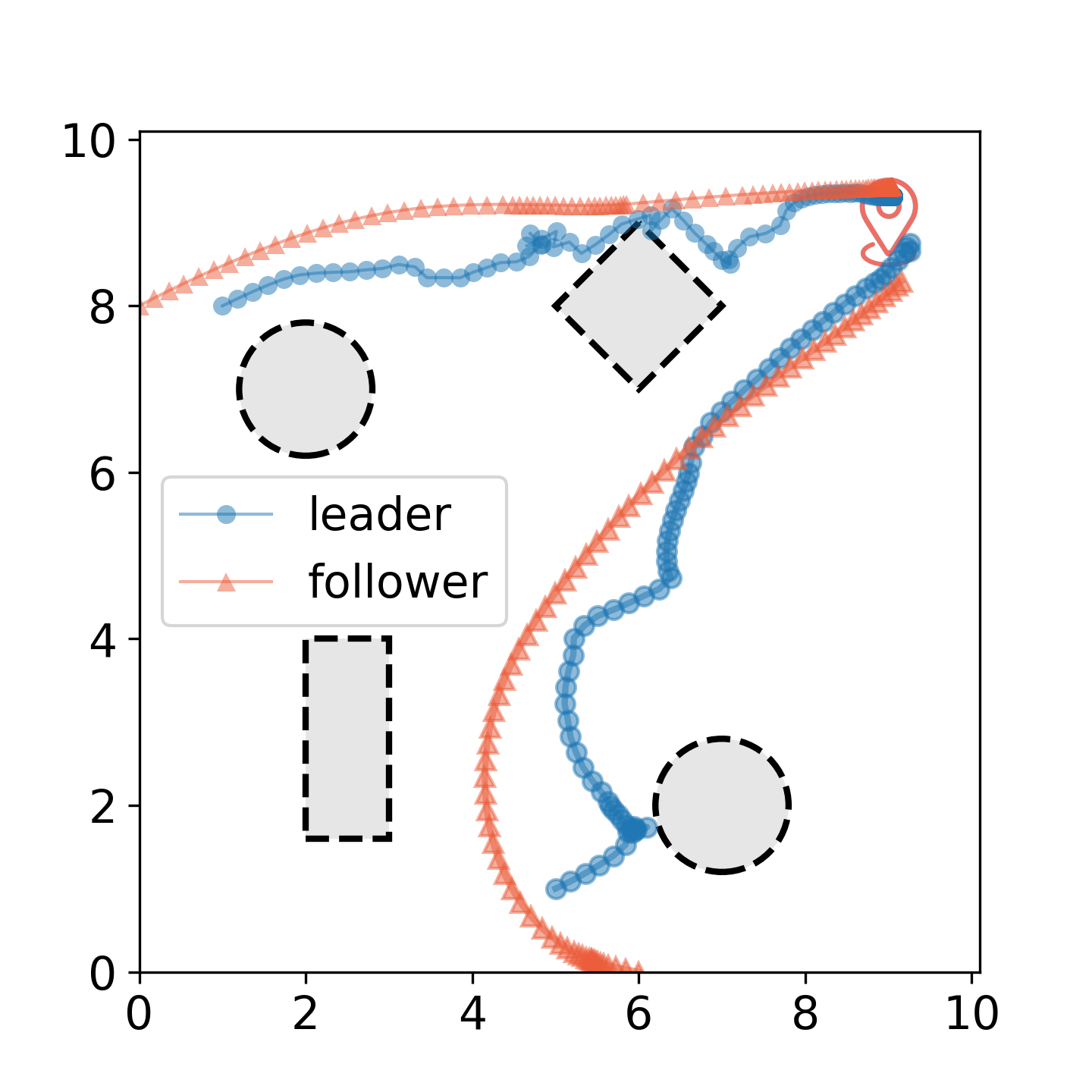}
        \caption{Type 3 follower.} 
        \label{fig:type.3}
    \end{subfigure}
    \captionsetup[subfigure]{justification=centering}
    \begin{subfigure}[b]{0.19\textwidth}
        \centering
        \includegraphics[height=3.5cm]{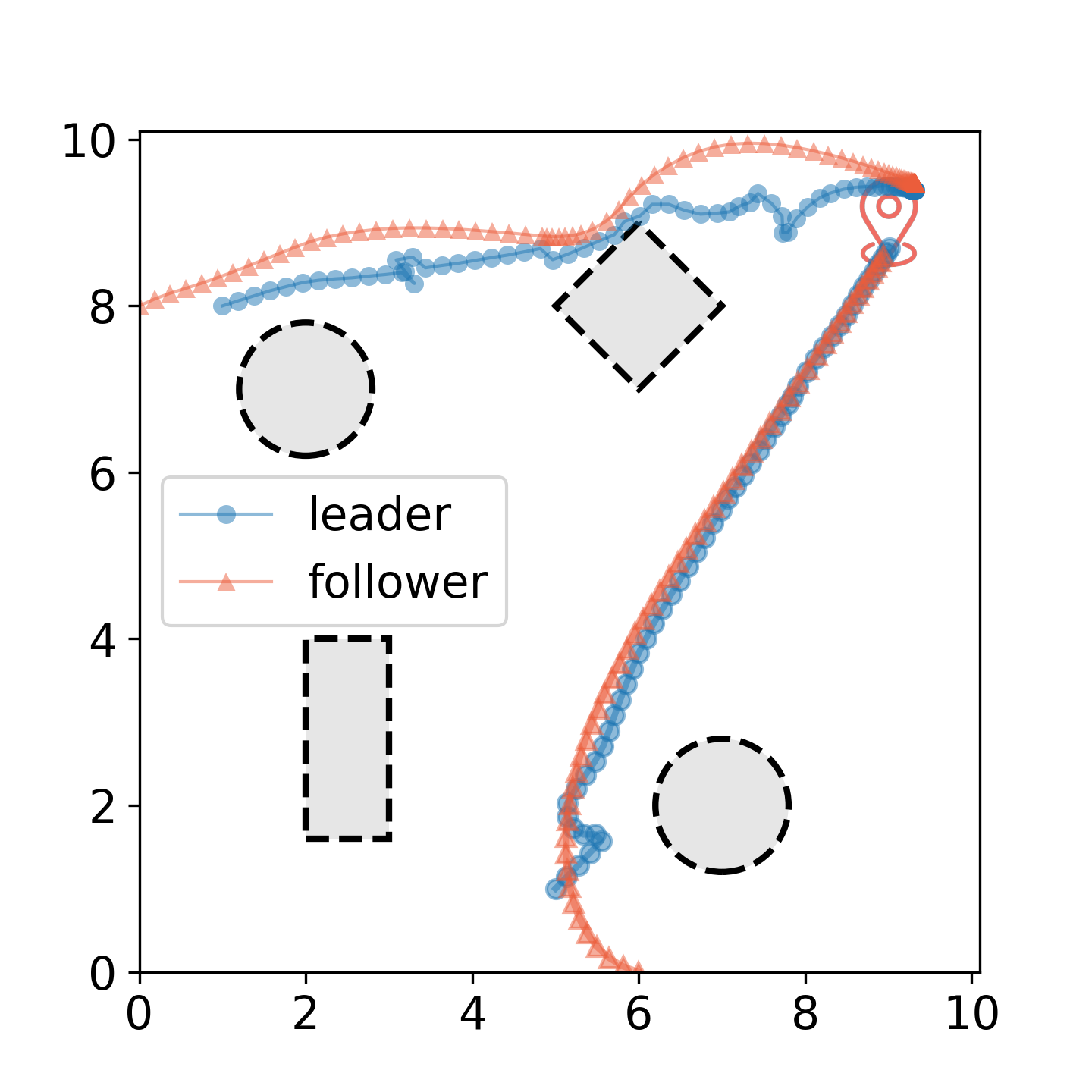}
        \caption{Type 4 follower.} 
        \label{fig:type.4}
    \end{subfigure}
    \captionsetup[subfigure]{justification=centering}
    \begin{subfigure}[b]{0.19\textwidth}
        \centering
        \includegraphics[height=3.5cm]{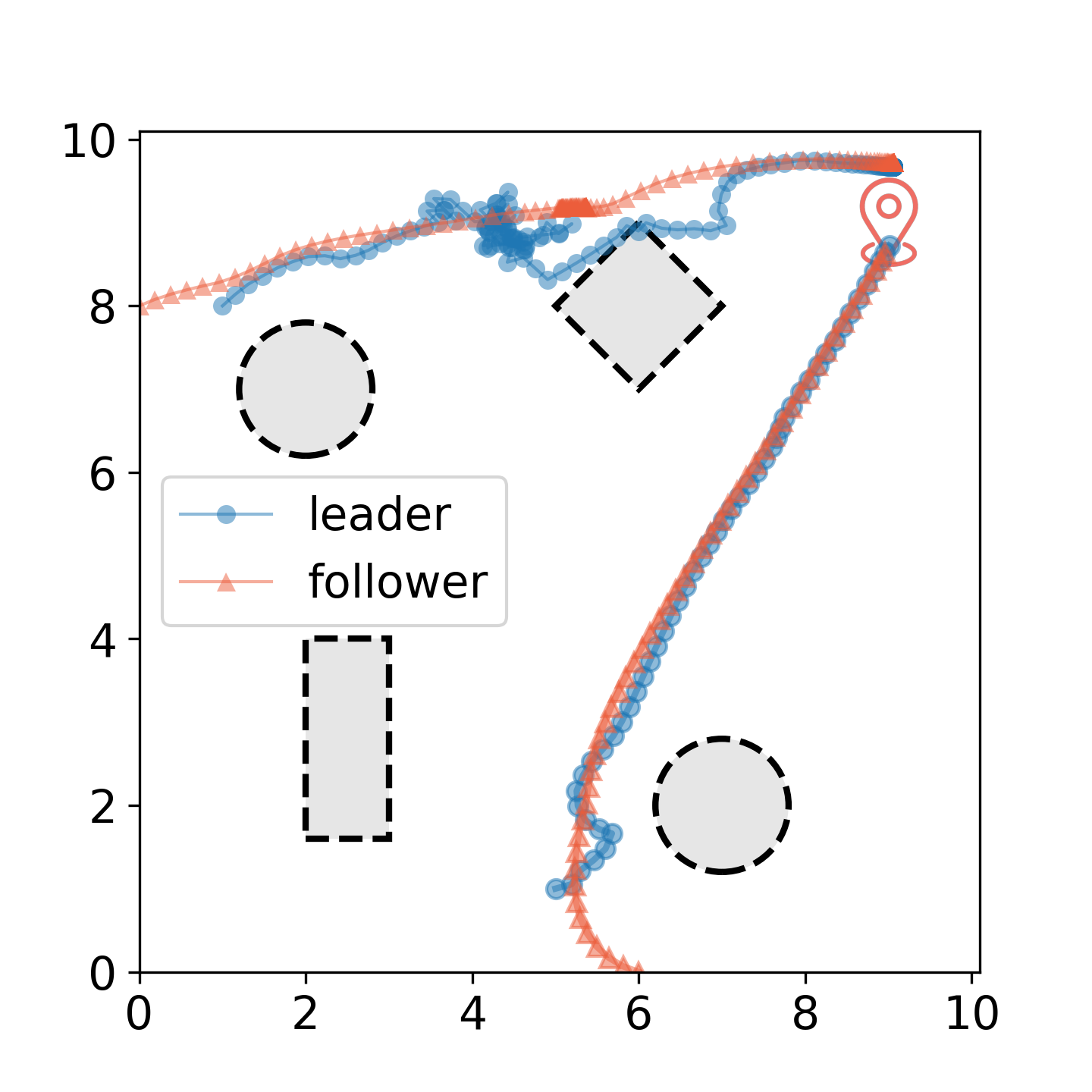}
        \caption{Type 5 follower.} 
        \label{fig:type.5}
    \end{subfigure}
    \caption{Guidance trajectories for different followers. The blue and the orange represent the leader and follower trajectories, respectively. Followers start from $[0,8]$ and $[6,0]$ to reach the goal region centered around $[9,9]$. The leader successfully guides all followers to the destination using adapted best-response models and receding horizon planning algorithms.}
    \label{fig:rc_trajectory}
\end{figure*}

After getting the adapted best-response model, the leader runs Alg.~\ref{alg:receding-horizon} to perform trajectory guidance. We simulate guided trajectories for each follower starting from different initial positions, shown in Fig.~\ref{fig:rc_trajectory}.

The leader successfully guides all types of followers to the destinations using corresponding adapted models, showing the effectiveness of the trajectory guidance. 
We can visualize different guidance plans that the leader uses for different followers. For example, we take the trajectory starting from $[6,0]$. As mentioned, type 2-3 followers tend to follow the leader more closely than type 4-5 followers. Therefore, the leader makes more aggressive trajectories in Fig.~\ref{fig:type.2}-\ref{fig:type.3} to attract the follower and help adjust their initial heading directions. While in Fig.~\ref{fig:type.4}-\ref{fig:type.5}, the leader simply needs to provide a reference trajectory to the follower because type 4-5 followers rely less on the guidance.

We also observe that the leader's trajectory is zigzagged near obstacles and smooth in flat regions. It shows that the leader is aware of the follower's behavior near the obstacle and adjusts her action to better guide the follower. Some trajectories can be complicated, such as in Fig.~\ref{fig:type.1} and Fig.~\ref{fig:type.5}. This is mainly due to the learning accuracy of the model. However, the leader still manages to guide the follower passing the obstacle, showing that the model is also effectively learned and robust.

\subsection{Comparison With Zero Guidance}

\begin{figure}
    \centering
    \begin{subfigure}[t]{0.23\textwidth}
        \centering
        \includegraphics[height=3.5cm]{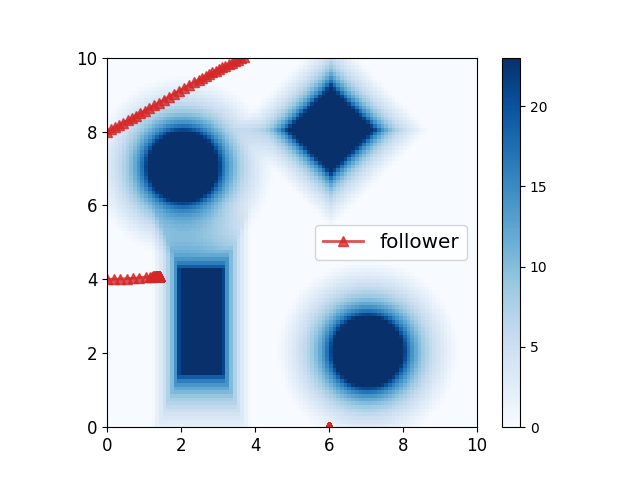}
        \caption{Type 3 follower.}
        \label{fig:noguide.1}
    \end{subfigure}
    \begin{subfigure}[t]{0.23\textwidth}
        \centering
        \includegraphics[height=3.5cm]{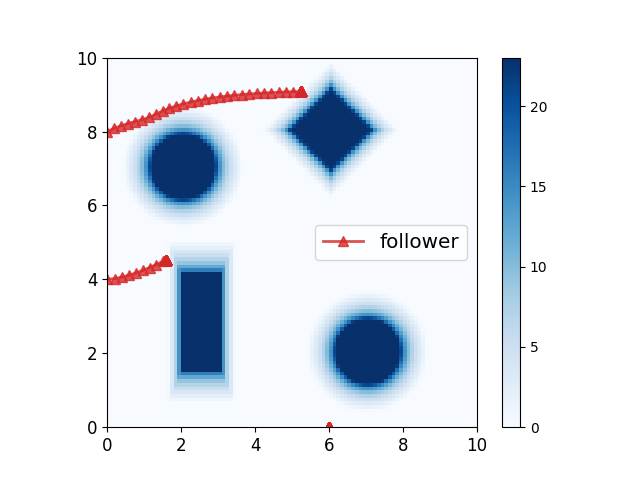}
        \caption{Type 5 follower.}
        \label{fig:noguide.2}
    \end{subfigure}
    \caption{Myopic trajectories for two types of followers starting from $[0,8]$, $[0,4]$, and $[6,0]$. None of them reach their destination. The color map represents the follower's sensing cost. We can see that type 3 follower has a wider sensing region than type 5 follower.}
    \label{fig:noguide}
\end{figure}

To demonstrate the necessity of the leader's guidance, we simulate a zero guidance scenario where the follower seeks a myopic trajectory by himself, i.e., disregarding the guidance cost in $J^F_\theta$. For simplicity, we plot the myopic trajectories for type $3$ and type $5$ followers starting from different positions, shown in Fig.~\ref{fig:noguide}. The myopic trajectories of type 1-3 followers are similar, while the ones of type 4-5 followers are similar. We also use color maps to plot the sensing cost for better visualization.

As we observe, two followers have trouble reaching their destination due to myopic planning. In fact, all followers fail to reach the destination from initial positions in the zero guidance scenario. They either get stuck on obstacles or run out of working space $\mX$, similar to type 3 and type 5 followers in Fig.~\ref{fig:noguide}. The reason for sticking at obstacles is twofold. First, followers use unicycle models, which do not have the flexibility to move around as freely as the leader (a single integrator). The second reason is related to rectangle obstacles. The follower senses the homogeneous cost along the obstacle surface. Thus, any action will lead to the same obstacle cost unless the follower can cross the obstacle in one step. Besides, taking action can also incur a control cost. So the best option is to stay still. We observe that both type 3 and type 5 followers starting from position $[0,4]$ and the type 5 follower starting from $[0,8]$ face the same issue. The guidance from the leader can effectively steer the follower away from obstacles and reach the destination (see Fig.~\ref{fig:rc_trajectory}), showing the value of trajectory guidance.

\section{Conclusion} \label{sec:conclusion}
In this work, we have proposed a Stackelberg meta-learning approach to address guided trajectory planning problems with multiple follower robots with unknown decision-making models. Our approach not only provides a game-theoretic characterization of the leader-follower type of interactions in trajectory guidance tasks but also develops an effective learning mechanism to learn and fast adapt to different guidance tasks to assist followers in reaching their destination.
Simulations have validated our approach in providing successful trajectory guidance and show better generalization and adaptation performance in learning followers' behavior models than other non-meta-learning-based approaches. Comparisons with zero guidance scenarios have demonstrated the value of guidance in assisting the follower robot to the destination.
For future work, we would investigate the impact of the response model properties on the planning results, such as smoothness, in more general Stackelberg frameworks. We would also generalize our approach to other application domains, such as human-robot interactions.

\appendix
\subsection{Discrete Dynamic Models} \label{app:dynamic_model}
We discretize the continuous dynamical system using $\dd t = 0.2$.
We write $x^L = [p^L_x, p^L_y]$ and $u^L = [v^L_x, v^L_y]$ as leader's state and control. The leader's discrete model is given by
\begin{equation}
\label{eq:discrete_dynamics_leader}
    x^L_{t+1} = x^L_t + u^L_t \cdot \dd t.
\end{equation}
Likewise, we write $x^F = [p^F_x, p^F_x, \phi^F]$ and $u^F = [v^F, \omega^F]$ as follower's state and control. We use mixed discretization for the follower, which yields
\begin{equation}
\label{eq:discrete_dynamics_follower}
    x^F_{t+1} = x^F_t + \begin{bmatrix} v^F_t \cos(\phi^F_{t+1}) \\ v^F_t  \sin(\phi^F_{t+1}) \\ \omega^F_t\end{bmatrix} \dd t.
\end{equation}
The follower first adjusts the heading angle and then applies the linear velocity in the discrete dynamic model. Here we assume that all followers have independent dynamics and are unaffected by the leader, a special case of the general form in \eqref{eq:sg.dynamics_follower}.

\subsection{Leader and Follower Cost Functions} \label{app:cost_fn}
The leader's cost is the same for all $\theta \in \Theta$, which is given by
\begin{gather*}
     g_\theta(x, u^L, u^F) = \norm{x-x^\dd}^2_{Q_1} + \norm{p^L - p^F}^2_{Q_2} + \norm{u^L}^2_{R},\\
     q_\theta(x) = \norm{x-x^\dd}^2_{Q_{f1}} + \norm{p^L - p^F}^2_{Q_{f2}},
\end{gather*}
where $x^\dd = [p^\dd, p^\dd, 0] \in \R^5$ is the aggregated target state; $Q_1, Q_2, R, Q_{f1}, Q_{f2} \succeq 0$ are weighting parameters of appropriate dimensions. The term $\norm{p^L - p^F}$ is interpreted as the guidance cost, producing a high value if the leader is far from the follower. In the simulation, we set $Q_1 = 2\times I_5, Q_2 = 5\times I_5, R = I_2$, the terminal cost parameters are set as $Q_{f1} = 5 \times Q_1, Q_{f2} = 5 \times Q_2$.

To simulate the interactive data, we construct the follower's \emph{ground truth} cost function using the following four base functions:
\begin{equation}
\label{eq:follower_cost}
\begin{split}
    J^F_\theta(u^F; x, u^L) = c_1 \norm{p^F_+ - p^\dd}^2_2 + c_2 \norm{p^L_+ - p^F_+}^2_2 \\ 
    + c_3 \norm{u^F}^2_2 + \sum_{j=1}^M h( c_4 (\norm{\Lambda(p^F_+ - p^O_j)}_l-d_j)),
\end{split}
\end{equation}
where $h(x) = 10\log(x)$ if $0 < x \leq 1$ and $0$ otherwise. The interpretation of the terms associated with $c_1$-$c_3$ is straightforward. $c_4$ relates the follower's sensing capabilities. A smaller $c_4$ indicates that the follower has a wider sensing range and can avoid obstacles earlier. $\Lambda \in \R^2$ is a scaling factor to define obstacles with different shapes. For example, we can choose a proper $\Lambda$ to define elliptic obstacles if $l = 2$.
Note that the follower takes the state $x$ and the leader's action $u^L$ as the parameter. His myopic decision is based on the one-step prediction $p^L_+$ and $p^F_+$ using the dynamics \eqref{eq:discrete_dynamics_leader}-\eqref{eq:discrete_dynamics_follower}. 
We assume that different followers have the same cost function structure but different parameters $c_1$-$c_4$. We use \eqref{eq:follower_cost} as the Oracle to generate the best-response data for learning.

\addtolength{\textheight}{-12cm}   


\bibliographystyle{IEEEtran}
\bibliography{IEEEabrv, mybib}

\begin{thebibliography}{10}
\providecommand{\url}[1]{#1}
\csname url@rmstyle\endcsname
\providecommand{\newblock}{\relax}
\providecommand{\bibinfo}[2]{#2}
\providecommand\BIBentrySTDinterwordspacing{\spaceskip=0pt\relax}
\providecommand\BIBentryALTinterwordstretchfactor{4}
\providecommand\BIBentryALTinterwordspacing{\spaceskip=\fontdimen2\font plus
\BIBentryALTinterwordstretchfactor\fontdimen3\font minus
  \fontdimen4\font\relax}
\providecommand\BIBforeignlanguage[2]{{%
\expandafter\ifx\csname l@#1\endcsname\relax
\typeout{** WARNING: IEEEtran.bst: No hyphenation pattern has been}%
\typeout{** loaded for the language `#1'. Using the pattern for}%
\typeout{** the default language instead.}%
\else
\language=\csname l@#1\endcsname
\fi
#2}}

\bibitem{panagou2015distributed}
D.~Panagou, D.~M. Stipanovi{\'c}, and P.~G. Voulgaris, ``Distributed
  coordination control for multi-robot networks using lyapunov-like barrier
  functions,'' \emph{IEEE Transactions on Automatic Control}, vol.~61, no.~3,
  pp. 617--632, 2015.

\bibitem{bibuli2012guidance}
M.~Bibuli, M.~Caccia, L.~Lapierre, and G.~Bruzzone, ``Guidance of unmanned
  surface vehicles: Experiments in vehicle following,'' \emph{IEEE Robotics \&
  Automation Magazine}, vol.~19, no.~3, pp. 92--102, 2012.

\bibitem{nikolaidis2017game}
S.~Nikolaidis, S.~Nath, A.~D. Procaccia, and S.~Srinivasa, ``Game-theoretic
  modeling of human adaptation in human-robot collaboration,'' in
  \emph{Proceedings of the 2017 ACM/IEEE international conference on
  human-robot interaction}, 2017, pp. 323--331.

\bibitem{bo2016human}
H.~Bo, D.~M. Mohan, M.~Azhar, K.~Sreekanth, and D.~Campolo, ``Human-robot
  collaboration for tooling path guidance,'' in \emph{2016 6th IEEE
  International Conference on Biomedical Robotics and Biomechatronics
  (BioRob)}.\hskip 1em plus 0.5em minus 0.4em\relax IEEE, 2016, pp. 1340--1345.

\bibitem{wang2016kinematic}
Z.~Wang and M.~Schwager, ``Kinematic multi-robot manipulation with no
  communication using force feedback,'' in \emph{2016 IEEE international
  conference on robotics and automation (ICRA)}.\hskip 1em plus 0.5em minus
  0.4em\relax IEEE, 2016, pp. 427--432.

\bibitem{machado2016multi}
T.~Machado, T.~Malheiro, S.~Monteiro, W.~Erlhagen, and E.~Bicho,
  ``Multi-constrained joint transportation tasks by teams of autonomous mobile
  robots using a dynamical systems approach,'' in \emph{2016 IEEE international
  conference on robotics and automation (ICRA)}.\hskip 1em plus 0.5em minus
  0.4em\relax IEEE, 2016, pp. 3111--3117.

\bibitem{kavraki1996probabilistic}
L.~E. Kavraki, P.~Svestka, J.-C. Latombe, and M.~H. Overmars, ``Probabilistic
  roadmaps for path planning in high-dimensional configuration spaces,''
  \emph{IEEE transactions on Robotics and Automation}, vol.~12, no.~4, pp.
  566--580, 1996.

\bibitem{lavalle2001randomized}
S.~M. LaValle and J.~J. Kuffner~Jr, ``Randomized kinodynamic planning,''
  \emph{The international journal of robotics research}, vol.~20, no.~5, pp.
  378--400, 2001.

\bibitem{karaman2011sampling}
S.~Karaman and E.~Frazzoli, ``Sampling-based algorithms for optimal motion
  planning,'' \emph{The international journal of robotics research}, vol.~30,
  no.~7, pp. 846--894, 2011.

\bibitem{wang2009fast}
Y.~Wang and S.~Boyd, ``Fast model predictive control using online
  optimization,'' \emph{IEEE Transactions on control systems technology},
  vol.~18, no.~2, pp. 267--278, 2009.

\bibitem{mohanan2018survey}
M.~Mohanan and A.~Salgoankar, ``A survey of robotic motion planning in dynamic
  environments,'' \emph{Robotics and Autonomous Systems}, vol. 100, pp.
  171--185, 2018.

\bibitem{hang2020human}
P.~Hang, C.~Lv, Y.~Xing, C.~Huang, and Z.~Hu, ``Human-like decision making for
  autonomous driving: A noncooperative game theoretic approach,'' \emph{IEEE
  Transactions on Intelligent Transportation Systems}, vol.~22, no.~4, pp.
  2076--2087, 2020.

\bibitem{wang2021game}
M.~Wang, Z.~Wang, J.~Talbot, J.~C. Gerdes, and M.~Schwager, ``Game-theoretic
  planning for self-driving cars in multivehicle competitive scenarios,''
  \emph{IEEE Transactions on Robotics}, vol.~37, no.~4, pp. 1313--1325, 2021.

\bibitem{turnwald2019human}
A.~Turnwald and D.~Wollherr, ``Human-like motion planning based on game
  theoretic decision making,'' \emph{International Journal of Social Robotics},
  vol.~11, no.~1, pp. 151--170, 2019.

\bibitem{zhu2014game}
M.~Zhu, M.~Otte, P.~Chaudhari, and E.~Frazzoli, ``Game theoretic controller
  synthesis for multi-robot motion planning part i: Trajectory based
  algorithms,'' in \emph{2014 IEEE International Conference on Robotics and
  Automation (ICRA)}.\hskip 1em plus 0.5em minus 0.4em\relax IEEE, 2014, pp.
  1646--1651.

\bibitem{wang2020game}
M.~Wang, N.~Mehr, A.~Gaidon, and M.~Schwager, ``Game-theoretic planning for
  risk-aware interactive agents,'' in \emph{2020 IEEE/RSJ International
  Conference on Intelligent Robots and Systems (IROS)}.\hskip 1em plus 0.5em
  minus 0.4em\relax IEEE, 2020, pp. 6998--7005.

\bibitem{bacsar1998dynamic}
T.~Ba{\c{s}}ar and G.~J. Olsder, \emph{Dynamic noncooperative game
  theory}.\hskip 1em plus 0.5em minus 0.4em\relax SIAM, 1998.

\bibitem{koh2020cooperative}
J.~J. Koh, G.~Ding, C.~Heckman, L.~Chen, and A.~Roncone, ``Cooperative control
  of mobile robots with stackelberg learning,'' in \emph{2020 IEEE/RSJ
  International Conference on Intelligent Robots and Systems (IROS)}.\hskip 1em
  plus 0.5em minus 0.4em\relax IEEE, 2020, pp. 7985--7992.

\bibitem{sadigh2016planning}
D.~Sadigh, S.~Sastry, S.~A. Seshia, and A.~D. Dragan, ``Planning for autonomous
  cars that leverage effects on human actions,'' in \emph{Robotics: Science and
  systems}, vol.~2.\hskip 1em plus 0.5em minus 0.4em\relax Ann Arbor, MI, USA,
  2016, pp. 1--9.

\bibitem{fisac2019hierarchical}
J.~F. Fisac, E.~Bronstein, E.~Stefansson, D.~Sadigh, S.~S. Sastry, and A.~D.
  Dragan, ``Hierarchical game-theoretic planning for autonomous vehicles,'' in
  \emph{2019 International conference on robotics and automation (ICRA)}.\hskip
  1em plus 0.5em minus 0.4em\relax IEEE, 2019, pp. 9590--9596.

\bibitem{zhao2022stackelberg}
Y.~Zhao, B.~Huang, J.~Yu, and Q.~Zhu, ``Stackelberg strategic guidance for
  heterogeneous robots collaboration,'' in \emph{2022 International Conference
  on Robotics and Automation (ICRA)}, 2022, pp. 4922--4928.

\bibitem{finn2017model}
C.~Finn, P.~Abbeel, and S.~Levine, ``Model-agnostic meta-learning for fast
  adaptation of deep networks,'' in \emph{International conference on machine
  learning}.\hskip 1em plus 0.5em minus 0.4em\relax PMLR, 2017, pp. 1126--1135.

\bibitem{Jia2022crmrl}
H.~Jia, Y.~Zhao, Y.~Zhai, B.~Ding, H.~Wang, and Q.~Wu, ``Crmrl: Collaborative
  relationship meta reinforcement learning for effectively adapting to type
  changes in multi-robotic system,'' \emph{IEEE Robotics and Automation
  Letters}, vol.~7, no.~4, pp. 11\,362--11\,369, 2022.

\bibitem{gao2019fast}
Y.~Gao, E.~Sibirtseva, G.~Castellano, and D.~Kragic, ``Fast adaptation with
  meta-reinforcement learning for trust modelling in human-robot interaction,''
  in \emph{2019 IEEE/RSJ International Conference on Intelligent Robots and
  Systems (IROS)}.\hskip 1em plus 0.5em minus 0.4em\relax IEEE, 2019, pp.
  305--312.

\bibitem{luipers2021concept}
D.~Luipers and A.~Richert, ``Concept of an intuitive human-robot-collaboration
  via motion tracking and augmented reality,'' in \emph{2021 IEEE International
  Conference on Artificial Intelligence and Computer Applications
  (ICAICA)}.\hskip 1em plus 0.5em minus 0.4em\relax IEEE, 2021, pp. 423--427.

\bibitem{xu2022meta}
S.~Xu and M.~Zhu, ``Meta value learning for fast policy-centric optimal motion
  planning,'' in \emph{Robotics science and systems}, 2022.

\bibitem{richards2022control}
S.~M. Richards, N.~Azizan, J.-J. Slotine, and M.~Pavone, ``Control-oriented
  meta-learning,'' \emph{arXiv preprint arXiv:2204.06716}, 2022.

\bibitem{paruchuri2008playing}
P.~Paruchuri, J.~P. Pearce, J.~Marecki, M.~Tambe, F.~Ordonez, and S.~Kraus,
  ``Playing games for security: An efficient exact algorithm for solving
  bayesian stackelberg games,'' in \emph{Proceedings of the 7th international
  joint conference on Autonomous agents and multiagent systems-Volume 2}, 2008,
  pp. 895--902.

\bibitem{lecun2015deep}
Y.~LeCun, Y.~Bengio, and G.~Hinton, ``Deep learning,'' \emph{nature}, vol. 521,
  no. 7553, pp. 436--444, 2015.

\end{thebibliography}

\end{document}